\documentclass[sigconf]{acmart}

\usepackage{hyperref}

\usepackage{tabularx}
\usepackage{multirow}
\usepackage[usestackEOL]{stackengine} 
\usepackage{cellspace}
\usepackage{fixltx2e}
\usepackage{pifont}

\usepackage{makecell}

\usepackage{graphicx}
\graphicspath{ {./images/} }
\usepackage{pifont}
\usepackage{balance}
\usepackage{enumitem}
\usepackage{makecell}

\usepackage{mathtools}
\usepackage{booktabs}

\newcommand{\todo}[1]{\textcolor{red}{#1}}
\newcommand{\note}[1]{\textcolor{brown}{#1}}

\newcommand{\hjscalebox}{\scalebox{1.0}[1.0]}

\newcommand{\seconv}{S2E$_{\text{conv}}$}
\newcommand{\seconvminus}{S2E$^{-}_{\text{conv}}$}
\newcommand{\miniskip}{\vspace*{-.9\baselineskip}}
\newcommand{\shrink}{\vspace*{-.9\baselineskip}}

\def\wast#1{\rlap{*}\textsubscript{#1}} 

\AtBeginDocument{%
  \providecommand\BibTeX{{%
    \normalfont B\kern-0.5em{\scshape i\kern-0.25em b}\kern-0.8em\TeX}}}

\copyrightyear{2022} 
\acmYear{2022} 
\setcopyright{rightsretained} 

\acmConference[CIKM '22]{Proceedings of the 31st ACM International Conference on Information and Knowledge Management}{October 17--21, 2022}{Atlanta, GA, USA}
\acmBooktitle{Proceedings of the 31st ACM Int'l Conference on Information and Knowledge Management (CIKM '22), Oct. 17--21, 2022, Atlanta, GA, USA}
\acmISBN{978-1-4503-9236-5/22/10}
\acmDOI{10.1145/3511808.3557667}

\usepackage{etoolbox}
\makeatletter
\patchcmd{\maketitle}{\@copyrightpermission}{
  \begin{minipage}{0.3\columnwidth}
    \href{http://creativecommons.org/licenses/by/4.0/}{\includegraphics[width=0.90\textwidth]{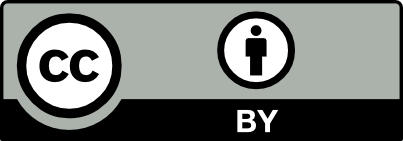}}
  \end{minipage}\hfill
  \begin{minipage}{0.7\columnwidth}
    \href{http://creativecommons.org/licenses/by/4.0/}{This work is licensed under a Creative Commons Attribution International 4.0 License.}
  \end{minipage}

  \vspace{5pt}
}{}{}

\makeatother

\begin{document}

\title{Personal Entity, Concept, and Named Entity Linking in Conversations} 


\author{Hideaki Joko}
\affiliation{%
  \institution{Radboud University}
  \country{}
}
\email{hideaki.joko@ru.nl}

\author{Faegheh Hasibi}
\affiliation{%
  \institution{Radboud University}
  \country{}
}
\email{f.hasibi@cs.ru.nl}


\renewcommand{\shortauthors}{Hideaki Joko \& Faegheh Hasibi}


\begin{abstract}
  Building conversational agents that can have natural and knowledge-grounded interactions with humans requires understanding user utterances. 
  Entity Linking (EL) is an effective technique for understanding natural language text and connecting it to external knowledge. It is, however, shown that the existing EL methods developed for annotating documents are suboptimal for conversations, where concepts and personal entities (e.g., ``my cars'') are essential for understanding user utterances.
  In this paper, we introduce a collection and a tool for entity linking in conversations. 
  We provide EL annotations for 1,327 conversational utterances, consisting of links to named entities, concepts, and personal entities. 
  The dataset is used for training our toolkit for conversational entity linking, CREL. Unlike existing EL methods, CREL is developed to identify both named entities and concepts. It also utilizes coreference resolution techniques to identify personal entities and their references to the explicit entity mentions in the conversations. We compare CREL with state-of-the-art techniques and show that it outperforms all existing baselines.

\end{abstract}



\maketitle

\section{Introduction}
\label{sec:introduction}

\begin{figure}[t]
  \includegraphics[width=1.0\linewidth]{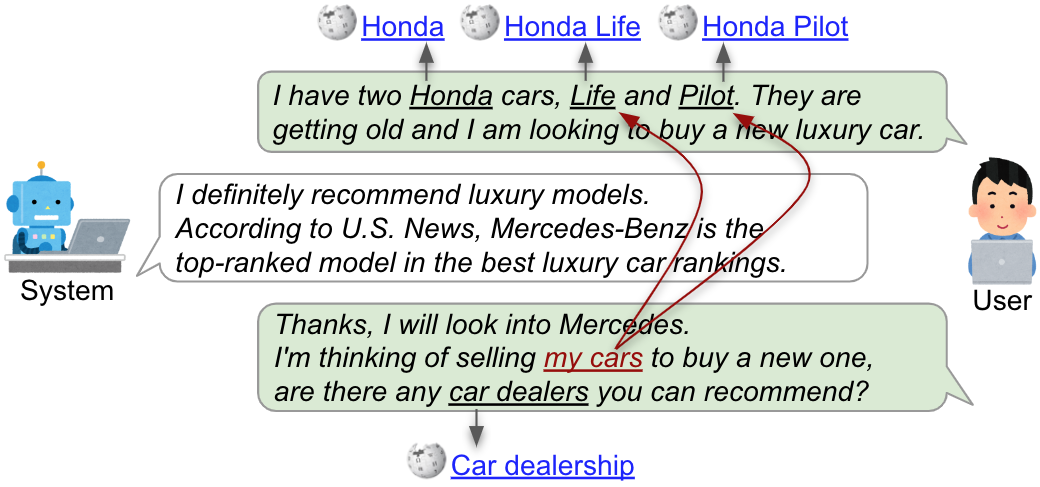}
    \caption{Example of entity linking in conversations.
      }
    \label{fig:example}
    \vspace{-1em}
\end{figure}

Understanding user utterances in conversational systems is fundamental to creating natural and informative interactions with users.
Entity Linking (EL) is a powerful text understanding technique that has been extensively explored in previous studies~\cite{Shang:2021:ERO, Joko:2021:CEL} and has been shown to be effective in question-answering and retrieval tasks~\cite{Yamada:2018:SOQ, Gerritse:2022:ETE, Dalton:2014:EQF, Hasibi:2016:EEL, Xiong:2017:WED, Dargahi:2018:QUE}. 
However, the EL approaches developed for documents prove to be suboptimal for conversations, to the extent that the state-of-the-art EL models are outperformed by the outdated models when evaluated on conversational EL datasets~\cite{Joko:2021:CEL}. The reasons are three-fold: 
 %
	\textbf{(i) Personal Entity Annotation}: Conversations often contain personal statements, referring to entities by their pronouns; e.g., ``my cars.'' Identifying personal entities is essential for building personal knowledge graphs~\cite{Balog:2019:PKG, Li:2014:PKG, Tigunova:2019:LBL, Tigunova:2020:CHA} and also generating personalized responses to users~\cite{Joshi:2017:PGD, Luo:2019:LPE}. Existing approaches for annotating documents, or even conversations~\cite{Laskar:2022:BEE, Xu:2022:OCR, Shang:2021:ERO}, do not identify and link personal entity mentions.
 \textbf{(ii) Concept Annotation}: EL models developed for documents or short texts focus mainly on annotating named entities~\cite{vanHulst:2020:REL, Cao:2021:GENRE}. In conversations, however, both named entities and concepts contribute to machine understanding of text~\cite{Joko:2021:CEL} and should be annotated.
 \textbf{(iii) Training Data}: Existing EL approaches are trained on well-written texts, while conversations have multiple (informal) turns with information spread over the turns.

In this paper, we overcome the above challenges and provide a toolkit and a training dataset as two resources for EL in conversations. The challenges are addressed in the following ways. 

\paragraph{Personal Entity Annotation}
We detect \emph{personal entity mentions} (e.g., ``my cars''), find their corresponding \emph{explicit entity mentions} (e.g., ``Life'' and ``Pilot''), and link them to the corresponding entities in the knowledge graph (see Figure~\ref{fig:example}).
This task and coreference resolution both connect and stand in contrast to each other: they both aim to find mentions referring to the same entity~\cite{Singh:2011:LCD, Lee:2018:HOC, Joshi:2019:BFC, Kirstain:2021:CRS}, but personal entity mentions are sparse and different from common pronominal expressions that are resolved in coreference resolution.
This implies that state-of-the-art coreference resolution methods cannot be used out-of-box for resolving personal entities. We, therefore, identify personal entity mentions in a separate module and modify coreference resolution techniques to map personal entities to their explicit entity mentions.
 
\paragraph{Concept Annotation} 
While open-domain knowledge graphs (e.g., Wikipedia) contain both named entities and concepts, conventional EL approaches are optimized for annotating only named entities, which hampers their performance for annotating conversations~\cite{vanHulst:2020:REL, Cao:2021:GENRE, Wu:2020:BLINK, Zhang:2022:ENT}. 
Specifically, the performance of named entity and concept linking in the widely used REL entity linker~\cite{vanHulst:2020:REL} varies dramatically (F-score of 43.1 and 2.5, respectively).
We mitigate this problem by training a mention detection model that can identify both concepts and named entity mentions in conversations and optimize the entity disambiguation model of REL~\cite{vanHulst:2020:REL} for all entity types. 
Our methods for personal entity, concept and named entity linking are integrated into our conversational EL tool, which is provided as a resource to the community. We show that this method outperforms state-of-the-art EL models.

\paragraph{Training Data}
A key to solving the aforementioned problems is training data. Existing studies on conversational entity linking either use private proprietary datasets~\cite{Shang:2021:ERO} or public datasets with insufficient number of conversations for both training and evaluation purposes~\cite{Joko:2021:CEL}.
In this work, we build on the existing ConEL dataset~\cite{Joko:2021:CEL} and address its shortcomings in our new dataset: (i) In ConEL, each personal entity mention is mapped to only a single explicit entity, 
while in real conversations, personal entity mentions can have multiple corresponding entities (see Figure~\ref{fig:example}); (ii) ConEL contains a limited number of conversations with personal entities, as it synthesizes conversations mainly from question-answering and task-oriented benchmarks with a limited number of personal statements. Additionally, the ConEL-PE dataset~\cite{Joko:2021:CEL}, generated by annotating real social chats, contains only 25 annotated conversations, which cannot be used for training purposes.
Our newly developed dataset, referred to as \emph{ConEL-2}, consists of 1,327 utterances, annotated with personal entities, concepts, and named entities. This dataset has been used for training and evaluation of our conversational EL tool.

In summary, this work makes the following resources available to the community:\footnote{\url{https://github.com/informagi/conversational-entity-linking-2022}}
\begin{itemize}[leftmargin=0.9em]
  \item An open-source tool for entity linking in conversations, which can uniquely identify personal entities, concepts, and named entities. Our developed conversational EL methods outperform state-of-the-art baselines.
  \item A dataset containing EL annotations for conversations, which can be used for training and evaluation purposes. 
\end{itemize}

\section{Method}
\label{sec:method}

Our conversational entity linking approach consists of two components; 
(1) personal entity linking, and (2) concept and named entity linking.


\subsection{Personal Entity Linking}
\label{sec:method_pel}
Inspired by coreference resolution approaches~\cite{Joshi:2019:BFC, Kirstain:2021:CRS, Lee:2017:E2E, Lee:2018:HOC}, our model identifies personal entity mentions and their entity antecedents from conversation history. We disentangle mention detection from entity antecedent assignment and employ a BERT-based model (cf. Sect.~\ref{sec:method_cne}) for detecting entity mentions and a rule-based approach~\cite{Li:2014:PKG} for identifying personal entity mentions. Specifically, we  follow \citet{Joko:2021:CEL} to identify personal mentions, starting with ``my'' or ``our'' tokens followed by adjectives, common nouns, proper nouns, pronouns, numbers, particles, and articles. Also, ``of'' and ``in'' are allowed to be part of the mention. 
While admittedly this approach cannot capture all forms of personal entity mentions, we believe that it covers a reasonably large number of cases to be used for early experimentation in this direction. We leave the comprehensive research on different forms of personal entity mention for future work.

%
%

Similar to~\cite{Kirstain:2021:CRS}, we denote the start ($s$) and end token ($e$) representations of mentions as:
\begin{gather}
  m^{s} = GeLU(\mathbf{W}^{s}v) \quad m^{e} = GeLU(\mathbf{W}^{e}v), 
\end{gather}
where $\mathbf{W}^{s}$ and $\mathbf{W}^{e}$ are trainable weights and $v$ is contextualized representation of a token.
We use LongFormer~\cite{Beltagy:2020:Longformer} to obtain contextualized embeddings as it can accommodate the long conversation history. The input to the LongFormer model is the current turn and conversational history. The hidden state in the output of the last layer of the model is used as $v$. 

Assuming mention $m_{bf}$ appears before mention $m_{af}$, the compatibility score between the two mentions is computed by:
%
\begin{align}
  score(m_{bf}, m_{af}) & =  m^{s}_{bf} \cdot B^{ss} \cdot m^{s}_{af}
        + m^{s}_{bf} \cdot B^{se} \cdot m^{e}_{af} \nonumber \\
        & + m^{e}_{bf} \cdot B^{es} \cdot m^{s}_{af}
        + m^{e}_{bf} \cdot B^{ee} \cdot m^{e}_{af}, 
 \label{eq:score}
\end{align}
where $B$ represents trainable weights and the superscripts $s$ and $e$ denote the start and end tokens of a mention.
Equation~\ref{eq:score} is computed for all pairs of personal entity mentions and explicit entity mentions.
During inference, personal and explicit entity mention pairs with the $score$ higher than threshold $\tau$ are selected.

\subsection{Concept and Named Entity Linking}
\label{sec:method_cne}
Entity linking typically involves two subtasks: (i) identifying mention boundaries and (ii) disambiguating mentions using a knowledge graph. 
In order to utilize existing advancements on these two subtasks, we need to adapt them to the conversational setup.
For mention detection (MD), this translates to adapting MD models to the conversational domain and identifying mentions of both named entities and concepts. 
We employ the vanilla BERT~\cite{Devlin:2019:BER} architecture and fine-tune it to predict BIO labels, corresponding to begin, inside, and outside of mentions. The BERT model utilized in this work is pre-trained on conversational datasets such as DailyDialogues~\cite{Li:2017:DDM}, OpenSubtitles~\cite{Lison:2016:OSE}, Debates~\cite{Zhang:2016:CFO}, and Blogs~\cite{Schler:2006:EAG}.

For entity disambiguation, we train REL~\cite{vanHulst:2020:REL} disambiguation approach on our ConEL-2 dataset.
REL is one of the state-of-the-art EL tools that is optimized for document annotation, but its disambiguation approach, considering both the local and global context of mentions, can be seamlessly adapted to conversations. Our entity linking method, consisting of the aforementioned MD and ED models is named CREL, which is an extension of REL tool for conversations.

\section{The ConEL-2 Collection}
\label{sec:annot}

To Facilitate the development of EL approaches for conversations, we construct the ConEL-2 collection.
We draw the conversations from an existing conversational dataset. To select the dataset, we focus on social chat datasets, as personal entities are mainly found in social chat conversations~\cite{Joko:2021:CEL}.
We choose the Wizard of Wikipedia (WoW)~\cite{Dinan:2018:WWK} dataset, which contains multi-domain multi-turn conversations collected from actual interactions between two people. The selected dialogues are annotated in three stages: (i) identifying personal entity mentions, (ii) pairing personal and explicit entity mentions, and (iii) entity linking. Unless stated otherwise, each HIT is annotated by three turkers, and additional turkers are added when needed based on our quality check strategies.

%


\begin{figure}[t]
  \includegraphics[width=1.0\linewidth]{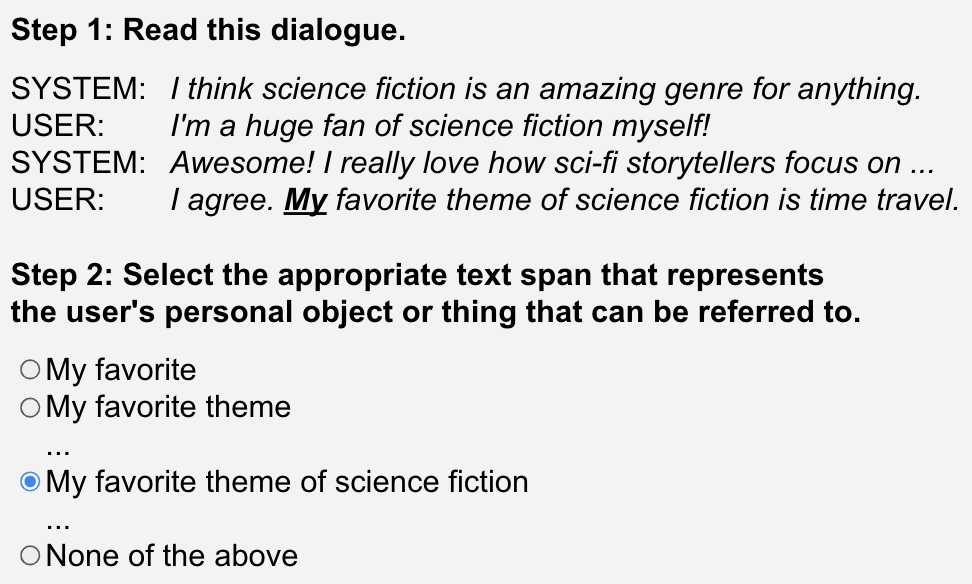}
    \caption{Annotation interface for identifying personal entity mentions (stage 1 of annotation process).}
    \label{fig:example_stage1}
\end{figure}

\medskip \noindent \emph{\textbf{Stage 1: Identifying personal entity mentions.}}
In this stage, we ask turkers to identify text spans referring to personal entities (i.e., personal entity mentions).
First, we extract conversations that contain ``my'' or ``our'' in at least one user utterance, which amounts to 7,888 out of 22,311 conversations in the WoW dataset.
To make the annotation process technically and financially feasible, we randomly select 1,000 conversations from the extracted dialogues.
We then show the dialogue history and a set of candidate mentions to the turkers and ask them to find the best personal entity mention;  the interface is shown in Fig.~\ref{fig:example_stage1}. Candidate mentions are text spans starting with ``my'' or ``our,'' followed by a maximum of 10 subsequent words. Our statistics on the collected annotations show that 98.9\% of the HITs were agreed upon by two or more turkers, with a Fleiss' kappa of 0.864.
We move forward with the 985 conversations, where at least two turkers agreed on all personal entity mentions of the conversation.

\begin{table}[t]
  \centering
  \caption{Statistics of the ConEL-2 collection. 
    Personal entity annotations represent the number of annotations for personal entity mentions, with or without corresponding explicit entity mention in the dialogue.
    }
  \label{tbl:annotation_statistics}
  \begin{tabular}{@{~}l|@{~~}l|@{~~}l|@{~~}l}
    \hline
    & Train & Val & Test \\
    \Xhline{1.5pt}
    Conversations & 174 & 58 & 58 \\
    User utterances & 800 & 267 & 260 \\
    NE and concept annotations & 1428 & 523 & 452 \\
    Personal entity annotations & 268 & 89 & 73 \\
    \hline
  \end{tabular}
\end{table}

\medskip \noindent \emph{\textbf{Stage 2: Pairing personal and explicit entity mentions.}}
In this stage, we ask turkers to find the corresponding explicit entity mention for a given personal entity mention. For example, in Figure~\ref{fig:example}, the personal entity mention ``\emph{my cars}'' is shown to turkers, and they are asked to find the corresponding explicit entity mention ``\emph{Life}'' and ``\emph{Pilot}.''
We create a pool of candidate explicit entity mentions by using existing EL tools: TagMe~\cite{Ferragina:2010:TAG}, WAT~\cite{Piccinno:2014:FTM}, REL~\cite{vanHulst:2020:REL}, and GENRE~\cite{Cao:2021:GENRE}.
The detected mentions are manually processed and meaningless duplicates such as ``a restaurant'' and ``restaurant'' are resolved.
Turkers are then given the option to select one corresponding explicit entity mention or the  ``not in dialogue'' option if the corresponding entity mention is not found in the dialogue.
Fleiss' kappa for the inter-annotator agreement is 0.434, which is considered moderate. 
To improve the annotation quality, two extra annotations are collected for HITs where two turkers agreed on one option. We then select all conversations with equal or more than three turkers agreed on one option (either an explicit entity mention or the  ``not in dialogue'' option). This selection amounts to 290 conversations, which are passed to the next step.

We note that one of the shortcomings of the existing ConEL dataset is identifying only one explicit entity mention for each personal mention (cf. Section~\ref{sec:introduction}). 
In ConEL-2, we address this issue and ask turkers to find another corresponding explicit entity mention for all extracted conversations if any.



\medskip \noindent \emph{\textbf{Stage 3: Entity linking}}
In this stage, we ask turkers to select the corresponding Wikipedia article to the given entity mention, 
e.g., mapping the mention ``car dealers'' to the entity \emph{Car dealership} in Figure~\ref{fig:example}.
We first use the EL tools described in the stage 2 to create the pool of candidate mention-entity pairs. 
Noisy mentions such as ``please'' and ``I am'' are manually removed from the pool.
We also include top-$10$ Wikipedia search\footnote{\url{https://www.mediawiki.org/wiki/API:Main_page}} results using mentions as queries.
Turkers are then asked to select one of the candidate entities for each entity mention if any.

The statistics show 98.7\% of the HITs were agreed by two or more turkers (Fleiss' kappa of 0.839), which highlights the high quality of our annotation process.
We therefore include all annotation results which are agreed by at least two turkers.
For the remaining 1.3\% of the non-agreed HITs, we recollected two extra annotations for each and select the entity that is selected by the majority of turkers.
The annotation results are shown in Table~\ref{tbl:annotation_statistics}. We split the dataset to train, validation, and test set with the 60\%-20\%-20\% ratio.


\section{Results}
\label{sec:results}

\subsection{Personal Entity Linking}
\label{sec:results_pe}

\begin{table}[t]
    \addtolength{\tabcolsep}{-1.9pt}
    \centering
    \miniskip
    \caption{Personal entity linking results. Micro-averaged precision, recall, and F1 scores are computed.
    }
    \shrink
    \label{tbl:pe}
    \hjscalebox{
      \begin{tabular}{l|ccc|ccc|ccc}
        \hline
        \multicolumn{1}{Sl|}{} 
        & \multicolumn{3}{Sc|}{ConEL-2 \footnotesize{(Val)}} & \multicolumn{3}{Sc|}{ConEL-2 \footnotesize{(Test)}} & \multicolumn{3}{Sc}{ConEL-PE} \\
        & \multicolumn{1}{Sc}{P} & \multicolumn{1}{Sc}{R} & \multicolumn{1}{Sc|}{F}
        & \multicolumn{1}{Sc}{P} & \multicolumn{1}{Sc}{R} & \multicolumn{1}{Sc|}{F}
        & \multicolumn{1}{Sc}{P} & \multicolumn{1}{Sc}{R} & \multicolumn{1}{Sc}{F} \\
        \Xhline{2pt}
        S2E~\cite{Kirstain:2021:CRS} & .375 & .415 & .394 & .292 & .302 & .297 & .317 & .481 & .382 \\
		W2V~\cite{Joko:2021:CEL} & .324 & .508 & .395 & .458 & .429 & .443 & .436 & .630 & .515 \\ 
        S2E$^{-}_{\text{conv}}$ & .494 & \textbf{.615} & .548 & .450 & \textbf{.571} & .503 & .400 & \textbf{.741} & .519 \\
        S2E\textsubscript{conv}& \textbf{.714} & .538 & \textbf{.614} & \textbf{.673} & .524 & \textbf{.589} & \textbf{.613} & .704 & \textbf{.655} \\
        \hline
        \end{tabular}
        }
  \end{table} 

  \begin{table}[t]
  	\shrink
    \addtolength{\tabcolsep}{-1.4pt}
    \centering
    \caption{Concept and named entity linking results.
    F\textsubscript{MD} and F\textsubscript{EL} represent the micro-averaged F1 scores for mention detection and entity linking, respectively.}
    \miniskip
    \label{tbl:md}
    \hjscalebox{
    \begin{tabular}{l|c@{~}c|c@{~}c|cc cc}
      \hline
      & \multicolumn{2}{Sc|}{ConEL-2 \footnotesize{(Val)}} & \multicolumn{2}{Sc|}{ConEL-2 \footnotesize{(Test)}} & 
      \multicolumn{2}{Sc}{ConEL} \\
      & F\textsubscript{MD} & F\textsubscript{EL} & F\textsubscript{MD} & F\textsubscript{EL} & 
      F\textsubscript{MD} & F\textsubscript{EL} \\
      \Xhline{2pt}
      GENRE \cite{Cao:2021:GENRE} & .290 & .252 & .320 & .299  & .350 & .211 \\
	  REL \cite{vanHulst:2020:REL} & .304 & .244 & .279 & .231 & .462 & .245 \\
      TagMe \cite{Ferragina:2010:TAG} & .559 & .478 & .611 & .504 & .510 & .375 \\
      WAT \cite{Piccinno:2014:FTM} & .616 & .539 & .613 & .519 & .416 & .336 \\
      \hline
      REL*(BERT$_{\text{conv}}$) & \textbf{.748} & \textbf{.652} & .716 & .587  & .551 & .424 \\
      REL*(BERT$_{\text{WP-conv}}$) & .697 & .624 & .708 & .592 & .508 & .399 \\
      REL*(BERT$_{\text{NER-conv}}$) & .723 & .635 & .693 & .576 & .548 & .407 \\
      CREL & .742 & .651 & \textbf{.729} & \textbf{.597} & \textbf{.559} & \textbf{.429} \\
      \hline
      \end{tabular}
    }
    \end{table} 
    
    \begin{table}[t]
      \small\addtolength{\tabcolsep}{-2pt}
      \centering
      \caption{Overall entity linking results, annotating both personal entities (PE) and concepts and named entities (EL).}
      \label{tbl:overall}
      \hjscalebox{
        \begin{tabular}{l l|ccc|ccc|ccc}
          \hline
          \multicolumn{1}{Sl }{} & \multicolumn{1}{Sl |}{}
          & \multicolumn{3}{Sc|}{ConEL-2 \footnotesize{(Val)}} & \multicolumn{3}{Sc|}{ConEL-2 \footnotesize{(Test)}} & \multicolumn{3}{Sc}{ConEL-PE} \\
         EL & PE & \multicolumn{1}{Sc}{P} & \multicolumn{1}{Sc}{R} & \multicolumn{1}{Sc|}{F}
          & \multicolumn{1}{Sc}{P} & \multicolumn{1}{Sc}{R} & \multicolumn{1}{Sc|}{F}
          & \multicolumn{1}{Sc}{P} & \multicolumn{1}{Sc}{R} & \multicolumn{1}{Sc}{F} \\
          \Xhline{2pt}
          REL & W2V & .523 & .159 & .244 & .516 & .159 & .243 & .488 & .206 & \underline{.290} \\
          REL & S2E\textsubscript{conv} & .523 & .162 & .248 & .522 & .165 & \underline{.250} & .466 & .206 & .286 \\ 
          \hline
          GENRE & W2V & .468 & .180 & .260 & .569 & .223 & .320 & .537 & .256 & .347 \\
          GENRE & S2E\textsubscript{conv} & .589 & .169 & .263 & \textbf{.675} & .217 & \underline{.328} & \textbf{.646} & .256 & \underline{.367} \\ 
          \hline
          TagMe & W2V & .504 & .407 & .451 & .530 & .440 & .481 & .528 & .518 & .523 \\
          TagMe & S2E\textsubscript{conv} & .612 & .395 & .480 & .621 & .434 & \underline{.511} & .636 & .518 & \underline{.571} \\ 
          \hline
          WAT & W2V & .506 & .489 & .497 & .545 & .464 & .501 & .536 & .628 & .579 \\
          WAT & S2E\textsubscript{conv} & .585 & .490 & .534 & .582 & .464 & \underline{.516} & .571 & .648 & \underline{.607} \\ 
          \hline   
          CREL & W2V & .636 & .589 & .612 & .607 & .554 & .579 & .573 & .714 & .635 \\     
          CREL & S2E\textsubscript{conv} & \textbf{.712} & \textbf{.593} & \textbf{.647} & .634 & \textbf{.566} & \underline{\textbf{.598}} & .600 & \textbf{.724} & \underline{\textbf{.656}} \\
          \hline

      \end{tabular}
      }
    \end{table}

\paragraph{Experimental Setup}
We take the LongFormer-large model\footnote{\url{https://huggingface.co/allenai/longformer-large-4096}}~\cite{Beltagy:2020:Longformer} pre-trained on OntoNote Release 5.0~\cite{Hovy:2006:ON9}, and fine-tune it on the training set of the ConEL-2 dataset. This model is referred to as \seconv{}. 
We perform training with the batch size of 8 and learning rate of 1e-5 using AdamW optimizer, and the validation set of ConEL-2 is used for early stopping.
For evaluation, we use validation and test sets of the ConEL-2 datasets as well as the ConEL-PE dataset~\cite{Joko:2021:CEL}. ConEL-PE contains 25 dialogues from the WoW dataset. We only use personal entity annotations for this evaluation.

\paragraph{Baselines}
We compare our model with the Wikipedia2Vec-based method (W2V)~\cite{Joko:2021:CEL}, where antecedent assignment is based on similarity between Wikipedia2vec~\cite{Yamada:2020:W2V} embeddings of  entities and personal entity mentions. 
We also report on the official S2E  model\footnote{\url{https://github.com/yuvalkirstain/s2e-coref}}~\cite{Kirstain:2021:CRS} and its variation \seconvminus{}, where fine-tuning is performed only on our conversational dataset (without using OntoNote dataset).

\paragraph{Results}
We measured the performance of our rule-based personal entity mention detection approach and obtained micro-averaged precision, recall, and F1 of 0.908, 0.898, and 0.903, respectively. This indicates that identifying personal entity mentions can be handled effectively by a simple approach. Assigning entity antecedents, on the hand, is a challenging task.

Table~\ref{tbl:pe} shows the results for personal entity linking. We observe that \seconv{} fine-tuned on both OntoNote and conversational datasets outperforms all baselines. It also indicates that, fine-tuning on conversational data substantially improves personal entity linking performance (S2E vs. \seconv{} results). 

%
%
%
%

\subsection{Concept and Named Entity Linking}
\label{sec:results_cne}

\paragraph{Experimental Setup}
For mention detection, we fine-tune bert-base-cased-conversational\footnote{\url{https://huggingface.co/DeepPavlov/bert-base-cased-conversational}} on our training data with the batch size of 16, learning rate of 5e-5, and AdamW optimizer, with the validation set of ConEL-2 for early stopping. For ED, we train the REL ED model, following the same training procedure as~\citet{vanHulst:2020:REL}.
As input for these models, we use only current turns for efficiency; although using the entire conversation history has shown slightly better performance~\cite{Joko:2021:CEL}, it doesn't change our conclusion, as it mainly benefits REL, with marginal benefits for WAT and TagMe.
For evaluation, we use the ConEL-2 and ConEL~\cite{Joko:2021:CEL} datasets. ConEL consists of 100 conversations that are collected from three different conversational tasks (question answering, task-oriented, and social chat) and annotated with concepts and named entities.
Only concept and named entity annotations are used for this evaluation.

\paragraph{Baselines}
REL~\cite{vanHulst:2020:REL}, GENRE~\cite{Cao:2021:GENRE}, TagMe~\cite{Ferragina:2010:TAG}, and WAT~\cite{Piccinno:2014:FTM} are strong publicly available EL tools that are used as our baselines. 
Additionally, we train three BERT-based MD models:
	(i) \emph{BERT$_{\text{conv}}$}: The bert-base-uncased model\footnote{\url{https://huggingface.co/bert-base-uncased}}, fine-tuned on our training data,
	(ii) BERT$_{\text{WP-conv}}$: Similar to BERT$_{\text{conv}}$, but fine-tuned on the Wikipedia anchor links from~\cite{Cao:2021:GENRE} before fine-tuning on ConEL-2 training data, and
	(iii) BERT$_{\text{NER-conv}}$: The bert-base-NER model\footnote{\url{https://huggingface.co/dslim/bert-base-NER}}, fine-tuned on our training data. 
The fine-tuning procedures are the same as CREL. We combine these models with the REL ED model trained on conversations (denoted as REL*) and report on their results.

\paragraph{Results}
The results are reported in Table~\ref{tbl:md}.
F\textsubscript{MD} and F\textsubscript{EL} are the F1 scores for MD and EL, respectively.
The results show that the REL and GENRE, while being state-of-the-art EL toolkits, perform worse than TagMe and WAT. 
This is because REL and GENRE are intended to detect only NEs, while TagMe and WAT are able to identify concepts as well.
Comparing CREL with REL* methods, we observe that adaptation of BERT for conversational domain plays an important role in conversational entity linking, reflected by the highest score for ConEL-2 (test) and ConEL datasets. 

\subsection{Overall Performance}
\label{sec:results_overall}

Putting all the pieces together, we report on the overall performance of our model for linking personal entities, concepts and NEs.
For this experiment, we report only on the ConEL-2 and ConEL-PE datasets, as ConEL does not have annotations for personal entities. The results are shown in Table~\ref{tbl:overall}.
We notice at first glance that our proposed  personal entity linking extension (S2E\textsubscript{conv}) leads to improved F-scores compared to W2V in most cases. The results also show that our EL tool (CREL with S2E\textsubscript{conv}) achieves the highest F-score for all sets, reinforcing the importance of adapting EL approaches for conversations.

\section{Conclusion}

In this paper, we tackled the problem of entity linking in conversational settings, where not only named entities, but also concepts and personal entities are important.
To this end, we collected conversational entity annotations, which allows us to train and evaluate entity linking in conversations.
Additionally, based on the collected annotations, we develop a conversational EL tool.
Our empirical results show that our EL tool achieves the highest performance over conventional EL tools for documents and short texts.
The collected annotations and our EL tool with detailed instructions are publicly available as a research resource for further study in conversation.
Future work includes incorporating coreference resolution methods to identify more diverse personal entity mentions.

\balance
\bibliographystyle{ACM-Reference-Format}
\bibliography{references-base}

\end{document}